\documentclass[runningheads]{llncs}
\usepackage[T1]{fontenc}
\usepackage{graphicx}
\usepackage{subfig}

\usepackage{verbatimbox} 

\usepackage{subfiles} 

\usepackage{tcolorbox} 

\begin{document}


%
\title{NUTS, NARS, and Speech}
%
%
\author{Dwane van der Sluis\inst{} }

\authorrunning{D. van der Sluis}

\institute{
\email{ucabdv1@ucl.ac.uk}\\
\url{ } }
\maketitle              

\begin{abstract}

To investigate whether “Intelligence is the capacity of an information-processing system to adapt to its environment while operating with insufficient knowledge and resources”\cite{wang1995non},  we look at utilising the non axiomatic reasoning system (NARS) for speech recognition. This article presents NUTS: raNdom dimensionality redUction non axiomaTic reasoning few Shot learner for perception. NUTS consists of naive dimensionality reduction, some pre-processing, and then non axiomatic reasoning (NARS). With only 2 training examples NUTS performs similarly to the Whisper Tiny model for discrete word identification.

\end{abstract}

\begin{tcolorbox}
This is a post-peer-review version of this article. The final version is available online at: https://doi.org/10.1007/978-3-031-33469-6\_31 
\end{tcolorbox}

\section{Introduction}

`Artificial Intelligence' now covers a wide range of tasks such as image recognition, speech recognition, game playing, and protein folding, each of which can be performed at, near, or beyond human level. Over time the term has drifted in meaning, away from a ‘thinking machine’, toward systems that often can only be applied to a single task, do not improve without further training, and take large amounts of resources to train and run. For example, GPT-3, a large language model, is estimated to have cost over 4.6 million dollars to train\cite{lamdalabs}. These models can be opaque, difficult to interpret, and unable to explain why a particular prediction was made, or unable to provide any guarantees in failure scenarios. Predicate Logic, on the other hand, is capable of  robust and consistent decisions. One such predicate logic system, CYC\cite{lenat1990cyc}, aims to encode all common human knowledge in a knowledge graph. This means CYC has the limitations of predicate logic, one being that all axioms (in the knowledge graph) be true and consistent, otherwise false statements can be derived. Another approach is Non Axiomatic Reasoning. The Non Axiomatic Reasoning System (NARS) performs reasoning that does not assign an objective value of truth to a statement, but instead assigns a subjective value. This subjective value is not fully trusted, and is revised over time as new information arrives. NARS has the advantages that it 1) can cope with holding conflicting information in its knowledge base 2) can explain predictions, 3) requires less data for inference 4) explicitly implements logic choosing which concepts to remove rather than randomly forgetting.

While Open NARS for Applications (ONA) was designed with reasoning in mind, we choose to investigate its usefulness, and resource consumption, on perception, and in particular speech recognition. This is because as Peirce stated “abductive inference shades into perceptual judgement without any sharp line of demarcation between them.”\cite{peirce1903}, and advances in understanding one may shed light on the other. The integration of deep learning and logic reasoning is an open-research problem and it is considered to be the key for the development of real intelligent agents\cite{Marra2020}. We narrow the focus of this paper to the dimensionality reduction, and logic needed to convert auditory sensory data into category labels, and the resources required in the Open NARS for Applications (ONA) software platform. First we give background, from the recent discussion around the definition of intelligence, and then how our human nervous system fused from 2 independent systems, perhaps leading to different characteristics of it. We then give a limited literature review, and then explain our method and experiments. In the last section we give and discuss our results.  

\section{Background}

Until recently, and possibly still, industry (and maybe academia too) are interested in whether new tasks can be learnt by AI, and if so, can they be sold profitably to consumers. The developers are under no obligation to consider the environmental impact or safety concerns. That said, some do by choice, but there is little compulsion from a social or regulatory point of view. This is partly due to deep learning being “unreasonably good”\cite{sejnowski2020unreasonable} and partly due to no other known way of achieving the same level of performance.
A focus on resource consumption was created by adding it explicitly into the definition of artificial intelligence\cite{Wang2019workingDefn}. Invitations to comment on this definition produced much discussion\cite{wang2022}. In this discussion it was pointed out that industry has existing finite resource limitations\cite{LeggAReview}, which is true, however for most leader board tasks, resource consumption is not taken into account. It also appears that governments are hesitant to impose resource limitations on industry. 
A resource limitation is of interest as it prevents brute force approaches and opens up the possibility of investigating how fewer resources can be utilised over time. Brute force approaches can also encode an entire domain space, further limiting the conclusions that can be drawn. If a method requires fewer resources over time, as a task or operation is repeated, then this suggests a deeper (or perhaps more precise or over-fitted) understanding of that task or operation, which may be of interest in the investigation of intelligence. Wang's definition of intelligence\cite{Wang2019workingDefn} separates skill (e.g. playing chess) from intelligence, and contains an assumption of insufficient knowledge and resources (AIKR). Under this definition learning a new skill to the same level of ability as some other method but with fewer resources or an insufficient understanding of the world (i.e. imperfect knowledge) is advantageous. Thus intelligence and skill are separate concepts.

The focus of this paper is on the cognitive processes that underlie speech recognition. It is assumed that these are similar to the cognitive processes that underlie other forms of perception. However, as stated earlier, the line between perceptual judgement and abductive inference has no clear demarcation\cite{peirce1903}. This may be because the physiology underpinning these functions has different origins. Genetic patterning studies suggest that the ‘blastoporal nervous system’, which coordinates feeding movements and locomotion, and the ‘apical nervous system’, which controls general body physiology, evolved separately in our ancestors more than six hundred million years ago \cite{tosches2013bilaterian,arendt2016nerve}, and subsequently fused. This may help explain why we (humans) are still aware of differences in different parts of our nervous system, being aware of our cognition around feeding and locomotion, but have little to no awareness of our ‘apical nervous system’. There are arguments that perception and cognition are unified \cite{shanahan2005perception,chalmers1992high,goldstone1998reuniting,jarvilehto1998efferent,lakoff1999philosophy,barsalou1999perceptual,shimojo2001sensory}, and arguments against, that is, for modularity\cite{fodor1983modularity,prinz2006mind}. The debate between modularity and unified is beyond the scope of this paper. However, the important point for the purposes of this paper is that the mechanisms of perception are not fully available to us. We do not know, for example, how we recognise objects or how we understand speech. 
Speech is temporal in nature and involves nuanced differentiation between acoustically similar sounds, (for example b in bright, and f in fright). 

Model performance over the last few decades has steadily improved, however it is computationally expensive. Current state-of-the-art models rely on a low level acoustic model, followed by a language model. The acoustic model converts a sound wave into an encoded representation of a sound, and the language model gives the probabilities of the next word, given the last few words, along with the acoustic encoding. As said, language models, like GPT-3 require a large amount of data to train, which conflicts with how children learn a new word with very few examples\cite{bloom2002children}.
Wang’s definition of intelligence\cite{Wang2019workingDefn} is based on the idea that intelligence is about making the most of the resources that are available and that it is not always possible to know everything that is going on in the world. So, someone who is able to learn new skills quickly and efficiently, even if they don’t have a lot of knowledge about those skills initially, would be considered intelligent. One approach to understanding the mechanisms of intelligence is to consider the different ways in which it can be measured. One way to measure intelligence may be by looking at someone or some system’s ability to learn new skills, and then measure the quality of that skill, as well as the energy consumed to learn it and perform it once learnt. Speech recognition is one such potential skill. 

Generally large dimensionality reduction is needed to convert perceived inputs into symbols that logic can be applied on. One approach\cite{Blazek2021} is to cluster the inputs in the feature space before similarity and difference are calculated and used as input into a deep network that is trained. Another approach is to pre-process with a DL model specifically trained for that modality, i.e. YOLO for computer vision, and use the generated labels. The generated labels form a lower bound on the resolution of the logic system, i.e. if the labels are ‘dogs’ and ‘cats’, it would be difficult for a logic system to learn of a new breed of dog. If the logic system uses scalar output features of a DL model, e.g the bounding box of the cat, x by y pixels, and if only far away cats are seen, and then later a closer (and therefore much bigger) cat is seen, scaling issues can be created, as the system may not see the full range of sizes immediately, and needs to re-calibrate previous observations when the scale is readjusted on seeing a much larger, or smaller example. In speech these challenges are exhibited in the form of the dimensionality reduction from 16k samples per second to $\sim$2 words per second, with uneven speaking speed. Speech recognition has traditionally used labelled data sets which cost USD50/hr to hand label, limiting the training data set size into the 10,00 hours or less range. The resultant systems have low generalisability with many recent state of the art systems reporting <5\% word error rates, which collapses into the 30-40\% range when used on other, but similar datasets\cite{radford2022robust}. The exact costs of training models such as Whisper\cite{radford2022robust} with 1.6 billion parameters, are unknown, but the 175 billion parameter GPT-3 (109x larger), also by OpenAI, is estimated at 4.6 million USD. If the training costs were a constant multiplier of the number of parameters (they are not), the training of Whisper could be in the order of magnitude of 40k USD. The training data set of Whisper was 168k hours of 16KHz speech. This equates to a data set size of 19TB, approximately half the estimated size of GPT3 about 45TB of training data. The recent success of attention in other domains has also been applied to speech. Andrade et al.\cite{Andrade2018ANA} developed a 202K parameter neural attention model, we will refer to as ANAM, which also targeted at the Speech Command dataset. 

\section{Literature Review}

In the 1970’s speech recognition relied on hand crafted features. This changed as end to end differential systems were developed and new SOTA were reached\cite{radford2022robust}. These systems lack interpretability, while not important for speech recognition, are of interest if the features triggering decisions can be exposed and validated. With concept whitening\cite{Chen2020} it is possible to concentrate (grounded) meaning in single neurons to aid interpretability, but requires category labels, which may not be available at training time, and adds complication as category labels “need to address topics like the representation of concepts, the strength of membership in a category, mechanisms for forming new concepts and the relation between a concept and the outside world”\cite{Wang2006}. Another approach is Deep Logic Models which integrate deep learning and logic reasoning in an end-to-end differentiable architecture\cite{Marra2020}. This work leads onto Relational Reasoning Networks R2Ns\cite{Marra2021LearningRF} which perform relational reasoning in the latent space of a deep learner architecture. However they suffer an explosion in memory needed as the number of possible ground atoms grows polynomially on the arity of the considered relations. This underpins the useful implication of the AIKR. Shanaha \cite{shanahan2005perception} explores perception of objects via computer vision, with abduction, but does not describe how resource intensive the work was or if resource limits were reached. Johansson\cite{johanssonscientific} investigates learning match to sample and the relationships more, less and opposite. Noting the advantages that AGI has, allowing us to experiment with an agents internals, and giving the example where NARS layer 6 (variables for generalisation) is shown to be needed for the work Hammer\cite{hammer2019} does on a system designed to only process data from perception, i.e. has no predefined knowledge. Generalised identity matching, where a new example is matched to a reference sample has been shown to be possible in NARS\cite{johansson2022}, and further that the derived identity concept could generalise to novel situations. In unpublished work Durisek\cite{Durisek} postulated speech recognition leveraging phonemes may be possible with NARS. 

\section{Method}
For simplicity we attempt to identify single whole spoken words, which has been the focus of much research. We use a standard data set, the Speech Command v2\cite{warden2018speech}, which contains 35 single word commands, 0-9, back, forward and other confusing words (bird, bed). Each word had over 3000 recordings, each of 1 second in duration or less. As in Whisper\cite{radford2022robust}, we take 16 bit, 16KHz audio, on which 80 bin MEL (logarithmic) spectrum was calculated every 10 milliseconds. This produced 8000 (80x100) energy intensity values per second, which were normalised. Utterances shorter than 1 second were padded. This reduced the input dimension from 16000 to 8000, and is a standard pre-processing step in speech recognition. These 8000 values needed to be restructured to be passed into ONA. Data was presented to ONA in the form of Narsese\footnote{For Narsese see https://cis.temple.edu/$\sim$pwang/NARS-Intro.html} statements. As a simple example, we encode three examples as Narsese instances, A, B and C with $n$ properties each. The strength of the property relationship to the instance was encoded in the truth value, i.e. a property with a strength of 0.9 would be encoded as: 

\begin{equation} \label{eq:1}
<\{A\} \rightarrow [p1]>. \%0.9\% 
\end{equation}

meaning “ ‘\{A\}’ has the property ‘p1’ with strength of 0.9”. It was then asserted that \{A\} is a LABEL and \{B\} is a LABEL. e.g.  
\begin{equation} 
<\{A\} \rightarrow  LABEL >.
\end{equation}
And then the system was queried to see if C was labelled correctly: 
\begin{equation} 
< \{C\} \rightarrow LABEL>?
\end{equation}

After a grid search we set the number of labelled examples per class to 2 (+ unlabelled example = 3). We note this is the smallest number that allows similarity to be exploited.
With this setup and synthetic and real data we attempted to answer the following questions:
 
\begin{itemize}
    \item RQ1: Can non axiomatic reasoning, which can cope with conflicting information, be leveraged to perform speech recognition?   
    \item RQ2: Is there a computational or performance? 
\end{itemize}

For a baselines we used 1) general speech recognition pre-trained Whisper models, and 2) the earlier mentioned ANAM model. We expected these to produce state-of-the-art results, at the cost of larger computation. \\

\textbf{Experiment 1 NARS, Computational complexity} We expressed data in the same manner as above, encoding each real world utterance as 8000 properties, using the energy in each bin as the strength of a property as Narsese statements. Energy values below 0.5 posed a problem, as they expressed absence of a feature in the data, e.g. the word ‘Moo’, should not have high frequency ‘s’, or ‘t’ sound in it. To enable ONA to track the absence of something, we negate the property name, i.e.[mel16x9] becomes [NOTmel16x9], and subtract the truth value from 0.5, so that a low truth value 0.1, becomes 0.9. E.g. 
\begin{equation} 
<\{U_1\} \rightarrow [mel16x9]>. \%0.1\% 
\end{equation} 
is replaced with 
\begin{equation} 
<\{U_1\} \rightarrow [NOTmel16x9]>. \%0.9\% 
\end{equation} 
We took 3 random utterances of ‘one’ (from the 3893 possible), generated the 8000 values for each, then encoded these as Narsese statements. We then asserted $<\{U_1\} \rightarrow one>.$ For utterances 1 and 2, and then queried ONA to see if utterance 3 is similar to the label $<\{U_n\} \rightarrow one>?$. All performance tests used a 64GB AMD Ryzen 5 3600 6-Core Processor running Ubuntu (no GPU). \\

\textbf{Experiment 2 Nalifier, NARS, Synthetic Data}
We take the same method as experiment 1, but this time pass the statements into a python pre-processor, Nalifier.py\cite{hammer2016opennars}, that suppresses certain Narsese statements, and synthesises other Narsese statements, which are in turn passed into ONA. The Nalifier has several functions, if the statement received consists of an instance property statement e.g. $<{utterance_n} \rightarrow [property_p]>$. It collects all the properties for this new instance, all the properties for all other instances in its memory and starts comparing them to find the closest. If an instance is found that the current instance is similar to, it synthesises and emits new narsese. The new narsese is passed into ONA (or more specifically NAL, the executable of ONA), and interpreted. The success criteria is the same as experiment 1, we check to see if the unknown instance is labelled correctly. \\

\textbf{Experiment 3 - NUTS}
We now introduce `NUTS' : raNdom dimensionality redUction non axiomaTic reasoning few Shot learner for perception. NUTS consists of four modules, dimensionality reduction,  conversion into narsese, a narses preprocessor (the Nalifier), and open NARS for applications (ONA). We used a random projection without sparsity, to reduce dimensionality, specifically we pass the input 16k samples through MEL encoding, producing $8000$ values. These $8000$ values were multiplied by a randomly generated $8000 * D$ matrix, reducing the dimensions to $D$. These $D$ values were then used to generate narsese as before, which is passed into the Nalifer which filers and generates narsese, which is passed into ONA. Each class was tested in turn with the negative classes consisted of the remaining 34 words in the speech command dataset. The number of learning examples of each word could be varied, along with ONA's setting for the size of the AIKR, the size of the reduced dimensionality space\footnote{A grid search showed 4 dimensions was reasonable.}, and the number of repeats. The matrix used for reduction was re-generated before each run. Success was measured as the proportions of runs where the correct "is a" relationship is identified. 

\section{Results}

\textbf{Experiment 1 NARS, computational complexity} As mentioned, baselines were OpenAIs whisper model, and Andrade's et al's ANAN. Whisper was tested on 100 random utterances from each of the 35 words in the Standard Commands data set, comparisons were case insensitive excluding punctuation. As seen in Table \ref{table:experiments_1_2_3}, Whisper tiny model took an average of 0.8 sec per inference (including encoding) with a performance of 58\%. ONA was unable to accurately identify the unknown utterance as being similar to anything in memory. This may have been because the full structure of the speech was not exploited, but we wished to avoid manual feature engineering. Analysis of the derived statements showed that the instances were considered similar to the properties rather than the instances, this was unexpected. \\


\textbf{Experiment 2 Nalifier, NARS, synthetic data} The Nalifier took considerable time to execute, to load and 'train' 2 instances with 2000 properties each, took 95 minutes. To load, encode and perform inference on a new example took an additional 43 min. This version of the Nalifier contains a $O(n^2)$ algorithm which executes each time a new property was observed for an instance. After 3 instances, each with 2000 properties were added into NARS, (first pre-processed by the Nalifier), NARS successfully determined that instance C, the unlabelled instance, was similar to instance A. This showed that speech recognition is possible with NARS, and that the Nalifier is needed. We did not attempt 8000 properties, or measure accuracy due to execution time. \\

\textbf{Experiment 3 NUTS} We were surprised randomly reducing dimensions worked, even for small numbers of training examples. For these experiments the best performance was obtained with 4 dimensions, when the unknown class was labelled correctly 64\% of the time, compared to 2.8\% for random performance, see Table \ref{table:experiments_1_2_3}. This compared favourably with Whisper Tiny's 58\%\footnote{Whisper leverages language models greatly improving multiword performance.}, but far below the ANAM's state of the art 94\%. Training was label and compute efficient, only needing 2 training samples per class, and inference time was 0.02 sec (including encoding and dimensionality reduction), far below that experiment 2's 43 min, showing that dimensionaility reduction is the source of the computational efficiency. 

\begin{figure}[t]
    \centering
    \begin{minipage}{0.48\linewidth}
        \centering
        \includegraphics[width=\linewidth]{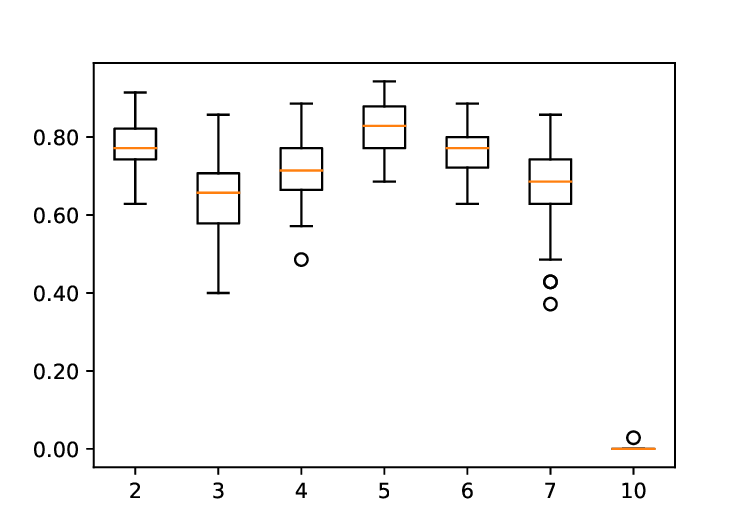} 
        \caption{Accuracy as a function of the reduced dimension embedding. Number of examples per class = 3.}
        \label{fig:c10a}
    \end{minipage}\hfill
    \begin{minipage}{0.45\linewidth}
        \centering
        \includegraphics[width=\linewidth]{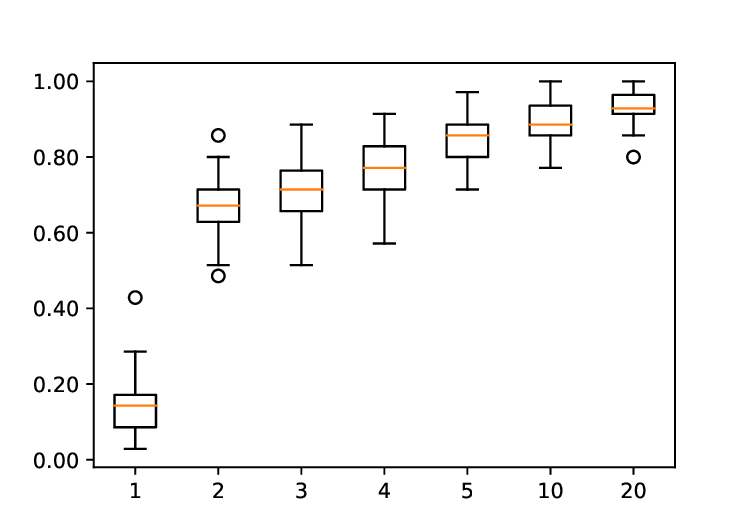} 
        \caption{Accuracy as a function of the number of examples. Reduced dimensions = 4.}
        \label{fig:c10b}
    \end{minipage}
\end{figure}

We conclude that perception, specifically speech recognition is possible with NARS. However we note performance collapsed certain words such as Bird, and Bed, yet ANAM's confusion matrix shows it is possible to distinguish them. This may be due to a limitation of NARS, or information loss in the dimensionaility reduction. Figure \ref{fig:c10a} shows the overall performance of a random generated matrix, a random word, and reduced dimensions 2–10, repeated 3500 times (100 times per class). Figure \ref{fig:c10b} shows performance increases with the number of examples, raising from 64\% at 2 examples to 90\% at 20.

shows the overall performance of a random generated matrix, and a random word, and reduced dimensions 2-5, repeated 3500 times (100 times per class). Best performance was obtained with 4 dimensions, with a mean of 59\%, and std 4\%, far above 2\% expected for random performance.  


\begin{table}[!h]
    \vspace{-5mm}
\caption{Performance \& Baselines: Whisper Tiny, Large, and ANAM}
	\centering
	\begin{tabular}{|l r r|r r|r r r r|}
	\cline{1-3} \cline{6-9} 
                                          & Exp1       &  Exp2     & & &   NUTS    &  Large    &  Tiny        &  ANAM \\ \cline{1-3} \cline{6-9} 
        Vocabulary Size                   &   1        &    2      & & &     35    & 50257     &   50257      &    35        \\ 
        Training Samples                  &   2        &    2      & & &   105     & 1e9 (est) &    1e9 (est) &    84843        \\ 
        Input Dimensions (1 sec audio)    &   16000    &   16000   & & &   16000   & 16000     &   16000      &    16000        \\ 
        Intermediate Dimensions           &   8000     &    200    & & &     4     & 8000      &   8000       &    9600        \\ 
        Inference Time (sec)              &  0.05      &  2615.00  & & &    0.02   &  43       &   0.80       &     0.095       \\ 
        Training Time (sec)               &   19       &    5700   & & &    16     &           &              &     7200        \\ 
        Performance  Accuracy             &  0\%      &        & & &    64\%   &  68\%     &   58\%       &     93\%      \\ \cline{1-3} \cline{6-9} 

    \end{tabular}
    \label{table:experiments_1_2_3}
    \vspace{-10mm}
\end{table}




\section{Discussion}


\textbf{RQ1}: Can non axiomatic reasoning, which can cope with conflicting information, be leveraged to perform speech recognition? Yes, we demonstrated that NARS (along with the Nalifier and dimensionality reduction), can interpret speech data in a meaningful way, obtaining 64\% accuracy with only 2 training examples on a 35 class problem. \\
\textbf{RQ2}: Is there a computational advantage? A crude comparison of inference time on CPU suggests they are in a similar order of magnitude. But as Whisper runs efficiently on GPU, and NUTS is a mix of C and Python, a strict comparison of counts of each operation type, could not be completed in the time, and it is left as further work. \\ 
%



We started by discussing the term `Artificial Intelligence' and resource consumption. While we did not produce a system with same or better performance with fewer resources, if we had done so would it be more `Intelligent'?  We would argue it would not, as the catalyst is not part of the described algorithm. We suspect building `intelligent' systems (as in `thinking machines') will remain elusive until the terms skill and intelligence are dis-entangled, and the catalyst for improvement is isolated and automated.  



\subsubsection{Acknowledgements} We thank reviewers and Parker Lamb for their comments.

\bibliographystyle{splncs04}

\newpage

\section{Appendix}

\subsection{Whisper Performance}

Table \ref{table:base_perf} gives the bench mark performance ignoring punctuation, whitespace and capitalization, for 51 random utterances out of each of the 35 classes. It should be noted that Whisper was designed to leverage language models, as such, its performance is usually much better than how we measured it here, on single words, with no context.

\begin{table}[!htb]
\caption{Base line performance}
	\centering
	\begin{tabular}{|l|l|l|l|l|l|}
	\hline
        model &   correct &  count  &  percentage correct & seconds/inference \\ \hline
        tiny &  1046 & 1785 & 0.58 & 0.8 \\ \hline
        small & 1267 & 1785 & 0.70 & 6.7 \\ \hline
        medium & 1296 & 1785 & 0.72 & 22.5 \\ \hline
        large & 1213 & 1785 & 0.68 & 43.3 \\ \hline
    \end{tabular}
    \label{table:base_perf}
\end{table}

\subsection{Individual Word performance}

Table \ref{table:word_perf} gives the average performance for each word across all dimensions, and average for 4 dimensions only, where we achieved the best overall performance.

\begin{table}[!htb]
\caption{Word performance, averaged across dimensions, and 4 dimensions only.}
	\centering
\label{table:word_perf}
\begin{tabular}{|l|r|r|c|l|r|r|c|l|r|r|} \cline{1-3} \cline{5-7} \cline{9-11}  
    Word &  Mean D1-10 &  Mean 4D & &    Word &  Mean D1-10  &  Mean D4 & &    Word &  Mean D1-10  & Mean D4 \\ \cline{1-3} \cline{5-7} \cline{9-11}  
     bed &  0.03 &   0.00 & &   seven &  0.12 &   0.31 & &   follow &  0.33 &   0.64 \\ \cline{1-3} \cline{5-7} \cline{9-11}  
     cat &  0.01 &   0.00 & &     tree &  0.17 &   0.40 & &     nine &  0.35 &   0.71 \\ \cline{1-3} \cline{5-7} \cline{9-11}  
    down &  0.00 &   0.01 & &       up &  0.18 &   0.44 & &   sheila &  0.37 &   0.84 \\ \cline{1-3} \cline{5-7} \cline{9-11}  
    five &  0.00 &   0.01 & & backward &  0.25 &   0.48 & &     stop &  0.38 &   0.90 \\ \cline{1-3} \cline{5-7} \cline{9-11}  
 forward &  0.00 &   0.01 & &   visual &  0.21 &   0.49 & &    right &  0.39 &   0.90 \\ \cline{1-3} \cline{5-7} \cline{9-11}  
      go &  0.00 &   0.01 & &    happy &  0.33 &   0.51 & &     zero &  0.39 &   0.90 \\ \cline{1-3} \cline{5-7} \cline{9-11}  
   house &  0.02 &   0.04 & &      dog &  0.27 &   0.57 & &    three &  0.39 &   0.91 \\ \cline{1-3} \cline{5-7} \cline{9-11}  
    left &  0.03 &   0.04 & &    learn &  0.33 &   0.57 & &      off &  0.36 &   0.91 \\ \cline{1-3} \cline{5-7} \cline{9-11}  
  marvin &  0.05 &   0.12 & &      yes &  0.22 &   0.59 & &      two &  0.39 &   0.93 \\ \cline{1-3} \cline{5-7} \cline{9-11}  
     six &  0.11 &   0.22 & &     bird &  0.28 &   0.59 & &      one &  0.39 &   0.94 \\ \cline{1-3} \cline{5-7} \cline{9-11}  
      no &  0.07 &   0.24 & &    eight &  0.31 &   0.60 & &      wow &  0.39 &   0.96 \\ \cline{1-3} \cline{5-7} \cline{9-11}  
      on &  0.08 &   0.27 & &     four &  0.33 &   0.63 & &          &       &        \\ \cline{1-3} \cline{5-7} \cline{9-11}  
\end{tabular}
\end{table}

\subsection{Speech Commands dataset v2}

List of words in the Speech Commands dataset v2: 'bed', 'cat', 'down', 'five', 'forward', 'go', 'house', 'left', 'marvin', 'no', 'on', 'seven', 'six', 'tree', 'up', 'visual', 'yes', 'backward', 'bird', 'dog', 'eight', 'follow', 'four', 'happy', 'learn', 'nine', 'off', 'one', 'right', 'sheila', 'stop', 'three', 'two', 'wow', 'zero'

\subsection{Additional Results }

\subsection{Narsese}

We had trouble encoding speech in a way that NARS could process it in a meaningful way. Many attempts were made. Here we list many of the formulations we attempted, and where remembered detail the type failure.
\begin{equation} \label{eq:appendix_1}
<A \rightarrow p1>. \%0.9\% 
\end{equation}
This form worked up to and including 10 properties on synthetic data, but failed on 11 and above properties.
\begin{equation} \label{eq:appendix_2}
<\{A\} \rightarrow [p1]>. \%0.9\% 
\end{equation}
This form worked on 500 properties on synthetic data, but failed at 2000 properties. The Nalifier took considerable time to execute, 90 minutes for 3 instances with 2000 properties each. The version of the Nalifier used at the time contained a $O(n^4)$ algorithm, which consumed most of the time. 8000 properties were not attempted.

\begin{equation} \label{eq:appendix_3}
<\{A\} \leftrightarrow {?1}>? 
\end{equation}
The above general is like form of the query failed.

\begin{equation} \label{eq:appendix_5}
<\{A\} \rightarrow One>? 
\end{equation}
This form of query worked (querying if the word was the word 'One'.

\begin{equation} \label{eq:appendix_n}
<\{A\} \rightarrow [NOTp1]>. \%0.9\% 
\end{equation}
This form of the was needed so that absence of signal could be used. The cut off of 0.5 was totally arbitrary, but seemed reasonable.

\subsection{Number of examples}

Inspired by triplet loss, and the fact that 2 examples are the minimum required for an example of similarity, we selected 2 training and 1 unknown samples as a starting point.

\subsection{Failure Analysis on individual words}

Considered very interesting, not completed due to time constraints.

\subsection{Dimensionality Reduction}

We tried sampling the MEL Spectrum to reduce input dimensions, this achieved around a 3\% success rate, very similar to random performance.

\subsection{Assumption of insufficient knowledge and resources}

\begin{figure}[!hb]
\centering
\includegraphics[width=0.5\textwidth]{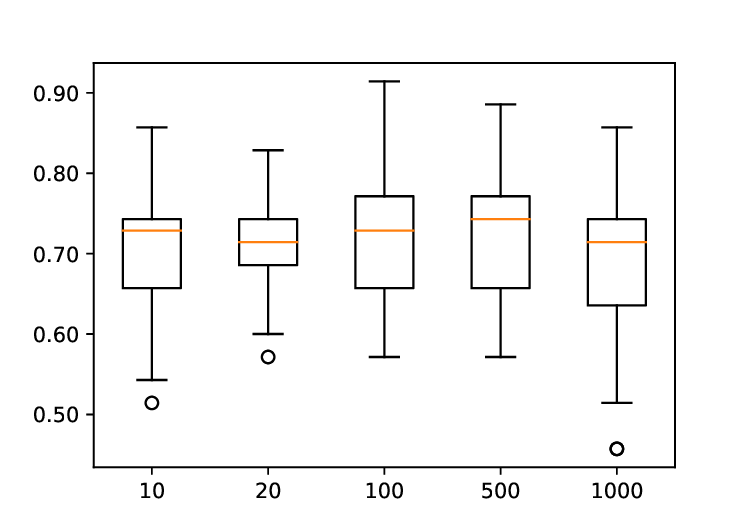} 
\caption{Performance as a function of the AIKR limit. Examples= 3, Dimensions=4 } \label{fig:perf_by_aikr}
\end{figure}

NARS contains a hyper-parameter, named AIKR, which specifies the amount of data (knowledge) that can be stored. Figure \ref{fig:perf_by_aikr} shows changing this hyper parameter had little impact on performance.

\vspace*{-1cm}

\subsection{Citation}

If you use this work, we would be pleased if you could cite it: \\
@inproceedings{10.1007/978-3-031-33469-6\_31,   \\
author = {van der Sluis, Dwane},   \\
title = {NUTS, NARS, and Speech}, \\
year = {2023}, \\
isbn = {978-3-031-33468-9}, \\
publisher = {Springer-Verlag}, \\
address = {Berlin, Heidelberg}, \\
url = {https://doi.org/10.1007/978-3-031-33469-6\_31}, \\
doi = {10.1007/978-3-031-33469-6\_31}, \\
booktitle = {Artificial General Intelligence: 16th International Conference, AGI 2023, Stockholm, Sweden, June 16–19, 2023, Proceedings}, \\
pages = {307–316}, \\
numpages = {10}, \\
keywords = {Non Axiomatic Reasoning, Perception, Few shot learning}, \\
location = {Stockholm, Sweden} \\
}

\end{document}